\documentclass{article}

\PassOptionsToPackage{numbers,compress}{natbib}
\usepackage[dblblindworkshop,final]{neurips_2026}
\workshoptitle{On-Device Intelligence: Foundation Models under Real-World Constraints}

\makeatletter
\renewcommand{\@noticestring}{NeurIPS 2026 Workshop on On-Device Intelligence: Foundation Models under Real-World Constraints.}
\makeatother

\usepackage[utf8]{inputenc}
\usepackage[T1]{fontenc}
\usepackage{hyperref}
\usepackage{url}
\usepackage{booktabs}
\usepackage{amsfonts}
\usepackage{amsmath}
\usepackage{nicefrac}
\usepackage{microtype}
\usepackage{xcolor}
\usepackage{graphicx}
\usepackage{multirow}
\usepackage{array}
\usepackage{xspace}
\usepackage{placeins}

\graphicspath{{figures/}}
\newcommand{\system}{\textsc{PhaseGate}\xspace}
\newcommand{\fixed}[1]{\textsc{Fixed}-#1}
\newcommand{\gate}[2]{\system $#1{\rightarrow}#2$}
\newcommand{\tpot}{\textsc{TPOT}\xspace}
\newcommand{\ttft}{\textsc{TTFT}\xspace}

\title{\texorpdfstring{\system}{PhaseGate}: Phase-Aware CPU Retrieval Scheduling for On-Device LLMs on Unified Memory}
\author{%
  Seoyoon Yum \\
  KAIST \\
  \texttt{gilbertyum@kaist.ac.kr}
  \And
  Sehoon Kim \\
  KAIST \\
  \texttt{sehoonkim@kaist.ac.kr}
}

\hypersetup{
  hidelinks,
  pdftitle={PhaseGate: Phase-Aware CPU Retrieval Scheduling for On-Device LLMs on Unified Memory},
  pdfauthor={Seoyoon Yum and Sehoon Kim},
  pdfsubject={NeurIPS 2026 On-Device Intelligence Workshop}
}

\begin{document}
\maketitle

\begin{abstract}
On-device assistants run GPU-based LLM inference alongside CPU retrieval on unified-memory systems.
Under a saturated local-retrieval workload, four concurrent retrieval workers raise 95th-percentile (p95) decode latency by 60--61\% on two M4 systems, whereas prefill latency rises by only 5.7--6.9\%.
We study LLM phase as an admission signal for independent CPU retrieval under controlled LLM workloads.
\system calibrates separate concurrency limits for prefill and decode, selecting four and one on our base-M4 configuration.
Under a backlogged queue, it achieves $2.0\times$ the aggregate retrieval throughput of the best tested feasible fixed policy, with both p95 LLM latency metrics within $1.25\times$ their no-retrieval baselines in all seven held-out runs.
A phase-blind control, TimeGate, uses the same two limits on a calibration-derived schedule without observing LLM phase.
It achieves similar retrieval throughput but violates the output-token latency limit in every run.
M2 and M2 Pro Mac minis reproduce the policy ordering, while output-length sweeps show that the advantage narrows as decode occupies more of each request.
\end{abstract}

\section{Introduction}

Compact models such as Apple's foundation model for Apple Intelligence and Google's Gemini Nano support on-device inference~\cite{gunter2024apple,geminiteam2023gemini}.
An on-device assistant also uses the CPU for retrieval, contextual-memory lookup, indexing, parsing, and tool execution.
Representative agentic applications spend substantial time in CPU-side tools~\cite{raj2025cpu}, and on-device vector-search systems must coordinate interactive queries with background ingestion and index maintenance~\cite{zhao2026muse}.

Server accelerators stream model state from GPU-local high-bandwidth memory (HBM), separate from host DRAM used by CPU tools~\cite{nvidia2022h100,zhong2024distserve}.
Apple Silicon gives the CPU and GPU access to a shared memory pool~\cite{apple2020m1}, making CPU pointer chasing and GPU weight streaming compete for memory service.
LLM inference has two phases: prefill processes prompt tokens in parallel and reuses weight loads, while decode emits one token at a time and repeatedly streams model state.

BWLOCK++ regulates CPU memory traffic during GPU execution using profiled bandwidth budgets~\cite{ali2018bwlock}.
FusionML splits prefill computation across the Apple-Silicon CPU and GPU while keeping decode on the GPU~\cite{mohite2026fusionml}.
We study a different control target: phase-dependent admission of independent CPU retrieval while keeping LLM execution on the GPU.

We use FAISS with a Hierarchical Navigable Small World (HNSW) index for approximate nearest-neighbor search~\cite{johnson2021faiss,malkov2020hnsw}.
HNSW performs irregular graph traversal.
Retrieval runs independently of the active LLM request; its queue stays backlogged during active periods to measure capacity under LLM latency limits.
On a fanless M4 MacBook Air, four HNSW workers increase p95 time-to-first-token (TTFT), our prefill-sensitive metric, by 5.7\%, but time-per-output-token (TPOT), our decode metric, by 61\%; a fan-cooled base-M4 Mac mini shows the same asymmetry.

\begin{figure}[t]
  \centering
  \includegraphics[width=\linewidth]{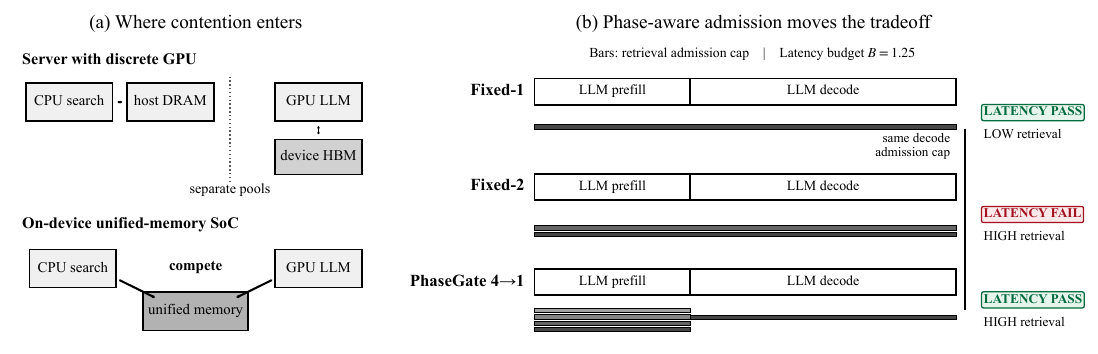}
  \caption{(a) A discrete-GPU server's CPU search and GPU LLM use separate memory pools, whereas Apple Silicon shares unified memory.
(b) On the measured base-M4 configuration, explicit PASS/FAIL labels show the tradeoff: Fixed-1 meets both latency limits but restricts retrieval, while Fixed-2 retrieves more but violates the output-token limit.
\system raises its admission cap during prefill and lowers it during decode; in-flight work may cross the boundary.
Schematic bars denote admission caps; PASS means both latency limits are met at $B=1.25$.}
  \label{fig:overview}
\end{figure}

A conventional retrieval pool uses one maximum concurrency throughout an LLM request.
We write Fixed-$n$ for a phase-oblivious concurrency cap of $n$, and $p\!\rightarrow\!d$ for \system's prefill and decode admission caps.
On the primary M4, Fixed-1 meets both latency limits, while the higher-throughput Fixed-2 violates output-token latency.
\system calibrates separate caps (Figure~\ref{fig:overview}), selecting 4$\rightarrow$1 on this configuration.
This pair is deployment-specific, and we compare it with a phase-blind timer that alternates the same limits without observing LLM phase.

\paragraph{Contributions.} (1) We quantify prefill--decode interference asymmetry using HNSW on two M4 systems and add a decode-only access-pattern control on the M4 Air.
(2) We evaluate a simple phase-based admission policy and isolate the value of phase alignment with a time-matched control.
(3) We evaluate the scheduler on M4, M2, and M2 Pro Minis, vary output length, and synthetically sweep the fraction of time with retrieval demand.\footnote{Code, experiment scripts, and processed results: \url{https://github.com/SeoyoonYum/PhaseGate}.}

\section{Observation: Phase-Asymmetric Interference}
\label{sec:observation}

\paragraph{Experimental Setup.}
We run 4-bit Qwen2.5-1.5B using MLX-LM v0.31.3~\cite{qwen2024,mlxlm2026} alongside a read-only FAISS HNSW index of 100,000 384-dimensional vectors with graph connectivity $M=32$.
On the M4 Mini, controlled synthetic traces use 2,048 input token IDs followed by 128 sequential single-token decode forward passes.
Tokenization, sampling, and natural-language output quality are outside this workload; Appendix~\ref{app:metrics} details the execution loop.
Retrieval uses two or four concurrent single-threaded FAISS HNSW calls, each processing 16 queries.
Concurrency caps count calls; throughput in queries per second (QPS) counts individual vector queries.

The queue remains nonempty during active periods, measuring aggregate retrieval capacity under an LLM-latency constraint.
Retrieval is independent of the concurrent generation, not prerequisite retrieval for that response.
Lower admission during decode can delay individual searches, especially for long outputs; we do not measure retrieval response time or end-to-end RAG latency.

\begin{figure}[t]
  \centering
  \includegraphics[width=\linewidth]{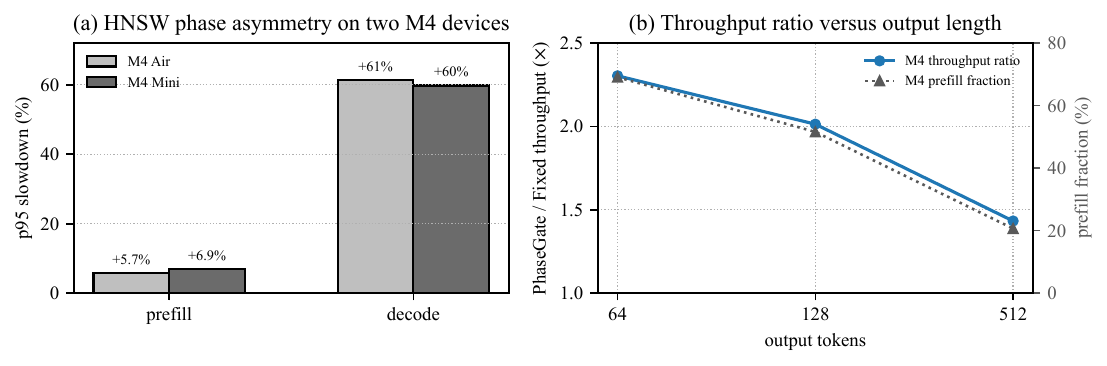}
\caption{(a) Four concurrent HNSW workers show the same p95 asymmetry on a fanless M4 Air and fan-cooled base-M4 Mini; each Mini bar is the median of three runs with 300 LLM requests per run.
(b) On the base-M4 Mini, \system/Fixed throughput falls with output length (five pairs per length); the right axis shows the median prefill fraction measured under \gate{4}{1}.}
  \label{fig:mechanism}
\end{figure}
\FloatBarrier

\paragraph{Decode Is More Sensitive to Concurrent Retrieval.}
On the M4 Mini, we first measure LLM latency without retrieval, then co-run either two or four concurrent HNSW calls.
Each concurrency setting uses three runs of 300 LLM requests, with the two settings evaluated in randomized order.
TTFT spans request start to the first completed decode step; TPOT summarizes subsequent step-to-step gaps.
With two concurrent calls on the M4 Mini, p95 TTFT increases by only 1.3\%, whereas p95 TPOT increases by 31\%.
With four, Figure~\ref{fig:mechanism}a shows 5.7\% versus 61\% on the Air and 6.9\% versus 60\% on the Mini; the Mini shows greater decode sensitivity in every repetition.
This pattern is consistent with prefill amortizing weight reads while decode has little reuse.

\paragraph{Access Pattern Matters Too.}
After establishing the phase asymmetry, we isolate access pattern by fixing the hardware to the M4 Air and examining decode only.
An address-jumping random gather reports just 2.4~GB/s of logical bandwidth but increases p95 decode latency by 29\%, whereas a contiguous 51~GB/s scan increases it by 27\%.
Thus byte rate alone is a poor interference proxy: irregular access can cause comparable slowdown to a much higher-rate sequential scan, which is relevant to graph-based retrieval such as HNSW~\cite{malkov2020hnsw}.

\section{\texorpdfstring{\system}{PhaseGate}}

\system uses the observed LLM phase to admit queued retrieval under separate prefill and decode caps.

\paragraph{Policy and Calibration.}
Let $p$ and $d$ be the prefill and decode concurrency limits.
A phase-oblivious scheduler uses one limit throughout (e.g., $p=d=1$); \system permits $p>d$.
The deployment supplies a baseline-relative budget $B$; $B=1.25$, for example, allows p95 TPOT and TTFT to rise by at most 25\% over their no-retrieval baselines.
Calibration tests candidate $(p,d)$ pairs and selects the pair with the highest retrieval QPS among those that consistently meet both limits.
Fixed caps assume conditions comparable to calibration, including background activity and memory pressure.
Phase labels do not measure current contention: compliance under the evaluated conditions is not a guarantee under changing load.
Material changes to hardware, model, sequence lengths, retrieval, or load require recalibration.

\paragraph{Runtime.}
Our instrumented MLX-LM harness emits prefill and decode events; \system admits a queued retrieval call only while the active-call count is below that phase's limit.
Calls already running are not preempted: when the phase limit decreases, they finish while new calls wait until the active count falls below the new limit.
The decode cap is thus an admission threshold, not an instantaneous bound on active work.
Effective isolation requires residual work to drain quickly relative to decode; jobs comparable to prefill duration are more likely to cross the boundary, and long or variable jobs can interfere well into decode.
We request no core affinity or thread-priority (QoS) override.

\paragraph{Evaluation.}
We compare \system with fixed-concurrency policies and a time-matched, phase-blind control, TimeGate, to measure the effect of phase alignment.
Appendix~\ref{sec:related} discusses related work.

\section{Experiments}

\paragraph{Experimental Setup.}
We use the same base-M4 Mini, model, HNSW index, and synthetic workload as in Section~\ref{sec:observation}.
We compare fixed caps, \system, and TimeGate under a backlogged retrieval queue.
For M4, we first compute the p95 inter-token gap within each request, then take the p95 of those values across requests.
Run-level p95 TTFT is computed directly across requests.
Appendix~\ref{app:experimental-details} gives setup details and metric definitions.

\subsection{Main Experiments}
\FloatBarrier

We calibrate \fixed{0}, \fixed{1}, \fixed{2}, \fixed{4}, and \gate{2}{0}, \gate{2}{1}, \gate{4}{0}, \gate{4}{1}, and \gate{4}{2} in three randomized 100-request runs per candidate.
A candidate is eligible only if both p95 latency metrics remain at or below the baseline-relative budget $B$ in every run.
Among tested budgets 10--45\% above baseline, $B=1.25$ is the smallest at which both a fixed policy with nonzero retrieval throughput and a \system policy are eligible.
We use it as a controlled comparison point, not a universal latency threshold.

\begin{table}[h]
  \caption{Base-M4 calibration sweep at $B=1.25$.
Throughput is the median of three runs; TPOT is the maximum per-run p95 ratio.
Unrounded ratios determine eligibility; $\dagger$ exceeds $B$.
All maximum TTFT ratios are $<1.1$; bold policies are selected.}
  \label{tab:calibration}
  \centering
  \small
  \setlength{\tabcolsep}{2.2pt}
  \begin{tabular}{
    >{\centering\arraybackslash}m{0.95cm}
    >{\centering\arraybackslash}m{2.35cm}
    >{\centering\arraybackslash}m{2.20cm}
    @{\hspace{7pt}\vrule width 0.45pt\hspace{7pt}}
    >{\centering\arraybackslash}m{1.85cm}
    >{\centering\arraybackslash}m{2.35cm}
    >{\centering\arraybackslash}m{2.20cm}
  }
    \toprule
    \shortstack{Fixed\\cap} & \shortstack{Retrieval\\throughput\\(kQPS) $\uparrow$} & \shortstack{Worst p95\\TPOT $\downarrow$\\($\times$ baseline)} & \shortstack{\system\\caps} & \shortstack{Retrieval\\throughput\\(kQPS) $\uparrow$} & \shortstack{Worst p95\\TPOT $\downarrow$\\($\times$ baseline)} \\
    \midrule
    0 & 0 & 1.021 & 2$\rightarrow$0 & 0.79 & 1.265$\dagger$ \\
    \textbf{1} & \textbf{0.78} & \textbf{1.247} & 2$\rightarrow$1 & 1.1 & 1.225 \\
    2 & 1.4 & 1.393$\dagger$ & 4$\rightarrow$0 & 1.2 & 1.271$\dagger$ \\
    4 & 2.1 & 1.690$\dagger$ & \textbf{4$\rightarrow$1} & \textbf{1.6} & \textbf{1.177} \\
      &     &     & 4$\rightarrow$2 & 1.8 & 1.384$\dagger$ \\
    \bottomrule
  \end{tabular}
\end{table}

Table~\ref{tab:calibration} exposes the central tradeoff.
Raising a constant cap from one to two or four increases retrieval throughput from 0.78 to 1.4 or 2.1~kQPS, but worst TPOT rises to $1.393\times$ or $1.690\times$ baseline, exceeding $B$.
\gate{4}{1} reaches 1.6~kQPS at $1.177\times$ worst TPOT; raising its decode cap to two reaches 1.8~kQPS but violates $B$ at $1.384\times$.
Calibration therefore selects \fixed{1} and \gate{4}{1}, the highest-throughput eligible policy in each family.
The zero-decode-cap policies have higher TPOT than the corresponding policies with a decode cap of one in all three runs.
The cause is unresolved; calibration evaluates each candidate directly without assuming monotonicity (Appendix~\ref{app:calibration-audit}).

We hold those choices fixed and evaluate them in seven paired randomized comparisons with 250 LLM requests per policy.
\fixed{1} and \gate{4}{1} meet both latency limits in all seven runs, but \gate{4}{1} reaches 1.6~kQPS versus 0.78~kQPS by raising the admission cap only during prefill (Table~\ref{tab:primary}).

\newpage
\subsection{Additional Analysis}

\paragraph{Phase-Blind Control.}
To test whether the throughput--latency tradeoff depends on alignment with LLM phase, we compare \system with a phase-blind control, TimeGate.
TimeGate replays a wall-clock schedule with the same high and low caps.
Its high-cap time fraction and switching rate match \system's calibration schedule; it does not observe runtime LLM phase.
TimeGate reaches a similar 1.5~kQPS, yet raises p95 TPOT by 60\% and meets both latency limits in 0/7 runs.
Its high-cap state overlaps 23\% of decode, compared with 0.003\% for \system; these are admission-state durations, not bounds on active CPU work.
This control supports the value of phase alignment in the tested design.

\begin{table}[h]
  \caption{Base-M4 evaluation at $B=1.25$.
Latency, throughput, and transition rows report medians across seven paired randomized repetitions with 250 requests per policy; ``7/7'' means both latency limits pass in all seven.}
  \label{tab:primary}
  \centering
  \small
  \setlength{\tabcolsep}{2.7pt}
  \begin{tabular}{
    >{\raggedright\arraybackslash}m{4.2cm}
    >{\centering\arraybackslash}m{2.1cm}
    >{\centering\arraybackslash}m{2.6cm}
    >{\centering\arraybackslash}m{2.4cm}
  }
    \toprule
    Metric & \fixed{1} & \gate{4}{1} & TimeGate 4/1 \\
    \midrule
    Runs meeting both limits & 7/7 & 7/7 & 0/7 \\
    p95 \tpot increase (\%) & 17 & 18 & 60 \\
    p95 \ttft increase (\%) & 0.61 & 4.6 & 4.3 \\
    Retrieval throughput (kQPS) & 0.78 & \textbf{1.6} & 1.5 \\
    p95 transition gap (ms) & 18 & 23 & 24 \\
    \bottomrule
  \end{tabular}
\end{table}
\FloatBarrier

\paragraph{Intermittent Retrieval Demand.}
We make retrieval active for 5\%, 25\%, or 100\% of wall-clock time, retaining a backlog during each active burst (five matched three-policy comparisons per duty; Appendix~\ref{app:duty}).
The median full-wall-clock \system/Fixed throughput ratio remains $2.0\times$ at all three retrieval duty cycles.
\fixed{1} and \system each meet both limits in all 15 runs, while TimeGate passes none; at 5\% duty, idle time lowers the absolute throughput to 79 versus 40~QPS without erasing the ratio.

\paragraph{Output-Length Sensitivity.}
We keep each device's selected caps fixed while varying output length to change the balance between prefill and decode.
On M4, \system/Fixed throughput falls from $2.3\times$ at 64 tokens to $2.0\times$ at 128 and $1.4\times$ at 512 as prefill's share of LLM processing time falls from 69\% to 52\% and 21\% (Figure~\ref{fig:mechanism}b).
At each M4 output length, both policies meet both $B=1.25$ limits in all five runs, using a no-retrieval baseline measured at that length.
The M2 Mini shows the same decreasing direction, from $1.8\times$ at 128 tokens to $1.4\times$ at 512 (see Appendix~\ref{app:m2-length}); devices are analyzed separately and the association is descriptive.

\paragraph{Cross-Device Replication.}
Table~\ref{tab:crossdevice} summarizes the device-specific comparisons.
M2 and M2 Pro summarize each request's token gaps by their mean, while M4 uses their p95.
The run-level p95 therefore tests different latency criteria on these devices (Appendix~\ref{app:metrics}).
The M2 Pro comparison uses a tighter budget and yields $2.5\times$, but includes neither a TimeGate comparison nor any pageout-free matched pairs, so it supports only the policy ordering.

\begin{table}[h]
  \caption{Mac mini evaluations, each comparing \fixed{1} and \gate{4}{1}. Throughput entries are median paired run ratios of \system to each control (two significant digits); passing runs meet both device-specific latency criteria (Appendix~\ref{app:metrics}), and ``--'' denotes an unevaluated control.}
  \label{tab:crossdevice}
  \centering
  \small
  \setlength{\tabcolsep}{1.5pt}
  \begin{tabular}{
    >{\raggedright\arraybackslash}m{2.15cm}
    >{\centering\arraybackslash}m{0.65cm}
    >{\centering\arraybackslash}m{2.45cm}
    >{\centering\arraybackslash}m{2.65cm}
    >{\centering\arraybackslash}m{1.65cm}
    >{\centering\arraybackslash}m{1.80cm}
    >{\centering\arraybackslash}m{1.80cm}
  }
    \toprule
    Device & $B$ & \shortstack{vs. Fixed\\throughput ($\times$)} & \shortstack{vs. TimeGate\\throughput ($\times$)} & \shortstack{Fixed\\passing runs} & \shortstack{\system\\passing runs} & \shortstack{TimeGate\\passing runs} \\
    \midrule
    Base-M4 Mini & 1.25 & 2.0 & 1.0 & 7/7 & 7/7 & 0/7 \\
    Base-M2 Mini & 1.25 & 1.8 & 1.0 & 5/5 & 5/5 & 0/5 \\
    M2 Pro Mini & 1.10 & 2.5 & -- & 7/7 & 7/7 & -- \\
    \bottomrule
  \end{tabular}
\end{table}
\FloatBarrier

\newpage
\subsection{Non-Preemptive Work and Transition Costs}
\label{sec:transition}

\paragraph{Retrieval That Continues into Decode.}
Calls admitted under the prefill cap of four may continue after the cap falls to one at decode entry.
We examine this work in seven held-out \gate{4}{1} traces, counting calls as active from admission to completion.

\emph{Drain time} measures the delay from decode entry until the active count falls to at most one and stays there for the rest of decode.
It is zero if the count never exceeds one during decode.
Table~\ref{tab:transition-audit} also reports the service time of each 16-query FAISS call, excluding queue waiting, and the delay until all calls active at decode entry finish.
Appendix~\ref{app:transition-audit} gives the trace-accounting details.

\begin{table}[h]
\centering
\caption{CPU service and drain times from seven held-out \gate{4}{1} traces (250 requests per run). The p95 column reports the median of seven per-run p95s; maxima span all seven traces.}
\label{tab:transition-audit}
\small
\begin{tabular}{lrr}
\toprule
Measure & \shortstack{Median of per-run\\p95s (ms)} & Observed maximum (ms) \\
\midrule
16-query call service time & 27.7 & 34.3 \\
Drain time to at most one active call & 26.6 & 31.5 \\
Completion of all calls active at decode entry & 29.0 & 32.2 \\
\bottomrule
\end{tabular}
\end{table}

The median per-run p95 drain of 26.6~ms and maximum of 31.5~ms are short relative to median phase durations of 1857~ms for prefill and 1739~ms for decode (medians of per-run medians).

Active calls exceed one during 1.06\% of decode time, although the requested cap is four during only 0.003\% (both medians across runs).
This difference reflects calls that continue after the admission cap falls.
The observed maxima characterize these short HNSW calls and do not bound longer or more variable workloads.
Such workloads, or shorter phases, require new calibration and service-time-tail measurements.
\FloatBarrier

\paragraph{GPU Transition and Memory Costs.}
We record token and phase timestamps in memory using a monotonic clock, without a monitoring process, to measure the prefill-to-first-token gap and two tail token-gap metrics.
The median paired increase in per-run p95 transition gap over Fixed-1 is 5.2~ms (0.29\% of baseline p95 TTFT).
The p99 inter-token gap and p95 per-request maximum token gap rise by less than 1\% (median paired changes).
No swap growth was observed, and memory pressure remained normal.
Small system-wide pageouts remain a potential confound.
The median paired throughput ratio remains $2.0\times$ in every leave-one-pair-out comparison.

\paragraph{Validity Checks.}
Fresh no-retrieval baselines vary by about 1.0\% within the continuous and duty-sweep M4 studies.
Swap growth, abnormal pressure, thermal warnings, corrupt traces, incorrect policy behavior, or severe drift invalidate a run, with one retry; small system-wide pageouts remain warnings.
All planned policy runs yield valid measurements, and invalid attempts remain in the artifact.

\section{Limitations and Conclusion}

\textbf{Limitations.}
The full protocol uses one base-M4 Mini; narrower evaluations use one each of M2 Mini, M2 Pro Mini, and M4 Air.
We evaluate one 1.5B 4-bit LLM with 2,048-token inputs; synthetic token traces do not test natural-language generation or sampling.
Scheduler and HNSW tests use read-only HNSW and device-specific caps.
The duty sweep uses backlogged bursts rather than random arrivals or full agent workflows.
Protocol and pageout differences prevent us from isolating cooling effects or ruling out effects from paging.
We do not evaluate long non-preemptive jobs, retrieval waiting-time tails, or online adjustment of the caps under changing background contention.

\textbf{Conclusion.}
On the base-M4 Mini, \system achieves $2.0\times$ the retrieval throughput of the best tested feasible fixed policy, meeting both LLM latency limits in all seven held-out runs.
The gain persists under intermittent demand and narrows as longer outputs increase the decode fraction.
TimeGate's similar throughput and greater decode slowdown support phase as a useful admission signal under these conditions.

\clearpage
{\small
\bibliographystyle{unsrtnat}
\bibliography{references}
}

\clearpage
\appendix
\setcounter{table}{0}
\renewcommand{\thetable}{A\arabic{table}}
\renewcommand{\theHtable}{appendix.\Alph{section}.\arabic{table}}
\setcounter{figure}{0}
\renewcommand{\thefigure}{A\arabic{figure}}
\renewcommand{\theHfigure}{appendix.\Alph{section}.\arabic{figure}}
\section{Experimental Details and Additional Results}
\label{app:experimental-details}

\subsection{Setup and Latency Metrics}
\label{app:metrics}

\begin{table}[h]
  \caption{Fan-cooled scheduler devices and shared setup (P/E CPU cores; UMA unified memory).}
  \label{tab:setup}
  \centering
  \footnotesize
  \setlength{\tabcolsep}{3.2pt}
  \begin{tabular}{
    >{\raggedright\arraybackslash}m{2.30cm}
    >{\raggedright\arraybackslash}m{11.1cm}
  }
    \toprule
    M4 primary & Mac mini (Mac16,10); 4P+6E CPU, 10-core GPU, 16~GB UMA \\
    M2 replication & Mac mini (Mac14,3); 4P+4E CPU, 10-core GPU, 16~GB UMA \\
    M2 Pro earlier & Mac mini (Mac14,12); 6P+4E CPU, 16-core GPU, 16~GB UMA \\
    M4 runtime & macOS 26.5/26.6.1 (continuous study/duty sweep); MLX 0.31.2; MLX-LM 0.31.3; FAISS 1.14.3; MLX memory limit 5.5~GB \\
    M4 model & \texttt{mlx-community/Qwen2.5-1.5B-Instruct-4bit}, revision \texttt{8b403126fc14f14cfc99bb4cfa72ecbc129ea677} \\
    Workload & 2,048 input tokens/128 decode steps; 100k 384-D vectors; queue kept nonempty \\
    \bottomrule
  \end{tabular}
\end{table}

\paragraph{Synthetic execution workload.}
The M4 Mini harness constructs deterministic input token IDs and prefills the transformer body to build the KV cache, omitting the vocabulary output head during prefill.
Each decode step evaluates the full model on a fixed single-token input, grows the KV cache, synchronizes execution, and records a timestamp.
It does not sample logits or feed generated tokens back into the model.
Here, output length counts these decode steps; ``token timestamps'' mark completed model evaluations.
This controls execution shape without measuring natural-language generation quality.

\paragraph{Latency metrics.}
Let $s_r$ be request $r$'s start time and $t_{r,i}$ its $i$th decode-step timestamp, recorded after model execution and synchronization.
TTFT is $t_{r,1}-s_r$; inter-token gaps are $t_{r,i}-t_{r,i-1}$ for $i\geq2$.
For the M4 Mini evaluations, run-level p95 TPOT is $Q_{.95,r}[Q_{.95,i}(t_{r,i}-t_{r,i-1})]$, where $Q_{.95}$ denotes the 95th percentile; p95 TTFT is $Q_{.95,r}(t_{r,1}-s_r)$.
M4 thus takes the p95 within each request and then across requests; the first-token boundary gap is reported separately.
The earlier M2 and M2 Pro protocols use the mean within each request and the p95 across requests.
They measure TTFT from recorded request arrival to the first token.
Each device uses a matching no-retrieval baseline.
These comparisons establish policy ordering within each protocol; their tail-latency criteria differ.

\FloatBarrier
\subsection{Calibration and Non-monotonicity}
\label{app:calibration-audit}

Table~\ref{tab:calibration-audit} provides the three randomized 100-request runs per candidate underlying Table~\ref{tab:calibration}.
All ratios use the same no-retrieval reference, the median of five baseline runs' p95 values: TPOT 12.046~ms and TTFT 1799.140~ms (rounded here; normalization uses full precision).
Throughput is the median across runs; feasibility requires both unrounded p95 ratios to be at most 1.25 in every run.
The maximum TTFT ratios are below 1.1 for every candidate, so TPOT determines eligibility here.

\begin{table}[h]
\centering
\caption{Per-repetition calibration results. Ratios are rounded to three decimals for display; eligibility uses unrounded values. Run indices are repetition identifiers, not a simultaneous paired comparison.}
\label{tab:calibration-audit}
\small
\setlength{\tabcolsep}{5pt}
\begin{tabular}{lrrrrrl}
\toprule
Policy & Median & \multicolumn{3}{c}{p95 TPOT / baseline} & Max TTFT & Eligible \\
 & kQPS & Run 0 & Run 1 & Run 2 & / baseline & at $B=1.25$ \\
\midrule
\fixed{0} & 0.000 & 1.009 & 0.996 & 1.021 & 0.980 & Yes \\
\fixed{1} & 0.784 & 1.213 & 1.235 & 1.247 & 1.008 & Yes \\
\fixed{2} & 1.379 & 1.325 & 1.294 & 1.393 & 1.016 & No \\
\fixed{4} & 2.136 & 1.667 & 1.690 & 1.667 & 1.074 & No \\
\gate{2}{0} & 0.793 & 1.261 & 1.265 & 1.258 & 1.002 & No \\
\gate{2}{1} & 1.146 & 1.218 & 1.157 & 1.225 & 1.016 & Yes \\
\gate{4}{0} & 1.218 & 1.257 & 1.271 & 1.251 & 1.027 & No \\
\gate{4}{1} & 1.573 & 1.177 & 1.170 & 1.172 & 1.042 & Yes \\
\gate{4}{2} & 1.789 & 1.289 & 1.384 & 1.352 & 1.054 & No \\
\bottomrule
\end{tabular}
\end{table}

The zero-decode-cap policies have higher TPOT than the corresponding policies with a decode cap of one in all three repetitions (Table~\ref{tab:calibration-audit}); rounding does not explain the ordering.
The cause remains unresolved, so calibration measures each candidate directly without assuming monotonicity or attributing the differences solely to noise.
Fixed-1's maximum ratio of 1.247 is close to the 1.25 threshold; its seven passing held-out runs support this configuration, not robustness to arbitrary background load.

\FloatBarrier
\subsection{M2 Output-Length Replication}
\label{app:m2-length}

The M2 Mini uses the same selected policies, with unchanged caps, at both output lengths.
The median \system/Fixed throughput ratio decreases from $1.8\times$ across five pairs at 128 output tokens to $1.4\times$ across three pairs at 512 tokens.
Both policies meet both $B=1.25$ limits in all five 128-token runs and all three 512-token runs, using length-matched no-retrieval baselines.

\begin{figure}[h]
  \centering
  \includegraphics[width=0.48\linewidth]{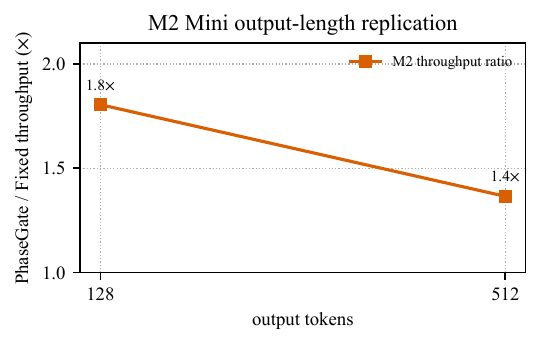}
  \caption{M2 Mini output-length replication with unchanged \gate{4}{1} and \fixed{1} caps.
The 128-token point summarizes five pairs and the 512-token point summarizes three; M2 is analyzed separately from M4.}
  \label{fig:m2-length}
\end{figure}

\FloatBarrier
\subsection{Intermittent Retrieval Demand}
\label{app:duty}
At 5\% and 25\% duty, 2.0-s backlogged bursts are separated by mean idle gaps of 38~s and 6.0~s, respectively.
Each duty, including continuous demand, uses five matched three-policy comparisons.
The 95\% paired-bootstrap intervals for the full-wall-clock \system/Fixed throughput ratio are [1.833, 2.039], [1.971, 2.052], and [1.987, 1.993], respectively (rounded to three decimals).
These experiments vary availability of queued work, not a distribution of individual request arrivals or deadlines.

\subsection{Trace Accounting and TimeGate Schedule}
\label{app:transition-audit}

Section~\ref{sec:transition} reports the CPU service and drain measurements.
The M4 calibration, mechanism, and scheduler evaluations use single-threaded FAISS calls with 16 vector queries per non-preemptive chunk, HNSW \texttt{efSearch}=128, and top-$k=10$.
A cap counts concurrent calls, and throughput counts completed vector queries.
The feeder groups 4,096 queries per task; its task-level latency field is not individual-query response time.

We verify the active-call count against every logged event.
Call service time excludes queue waiting; statistics include calls that start and finish between the first request start and last request completion.
Drain time in Table~\ref{tab:transition-audit} includes later increases within the same decode interval: the count must fall to at most one and remain there.
Completion of calls active at decode entry tracks only that initial set through its last completion; it is zero for an empty set.
For both statistics, we report the median of seven per-run p95s and the maximum across all traces.

\paragraph{TimeGate schedule.}
For the primary M4 evaluation, TimeGate cyclically replays 600 calibration-derived alternating high- and low-cap intervals (median durations 1.840/1.738~s; total period 1073.235~s).
Each repetition uses a uniformly sampled circular start offset, frozen before evaluation.
The wall-clock controller starts before retrieval warmup and runs across requests without resets or LLM-phase input.

\clearpage
\section{Related Work}
\label{sec:related}

\paragraph{Phase-Structured LLM Serving.}
Server systems exploit the different resource profiles of prefill and decode inside the inference stack.
Splitwise and DistServe assign the phases to separately provisioned GPU pools, Sarathi-Serve chunks prefills to avoid stalling ongoing decodes, and speculative decoding reduces sequential decode work~\cite{patel2024splitwise,zhong2024distserve,agrawal2024sarathi,leviathan2023speculative}.
\system is complementary: it leaves LLM execution unchanged on a single on-device GPU and uses the phase boundary to regulate an independent CPU retrieval workload.

\paragraph{Shared-Memory Interference.}
BWLOCK++ activates CPU memory-traffic throttling around GPU execution and profiles bandwidth budgets to protect integrated-GPU kernels~\cite{ali2018bwlock}.
Sereno detects mobile memory contention and yields background LLM inference to foreground applications~\cite{xin2026sereno}.
Adjacent GPU-sharing work such as REEF instead provides kernel preemption and controlled co-execution~\cite{han2022reef}.
FusionML distinguishes prefill from decode, partitioning prefill matrix operations across the Apple-Silicon CPU and GPU while leaving decode on the GPU~\cite{mohite2026fusionml}.
The control targets are CPU bandwidth budgets in BWLOCK++, LLM operator partitioning in FusionML, and admission of independent CPU retrieval in \system.
Our fixed-cap and TimeGate comparisons measure the throughput--latency effects of retrieval admission and phase alignment.

\paragraph{On-Device Retrieval.}
Apple's Foundation Models guidance presents retrieval-augmented generation as a way to fetch relevant context for an on-device model~\cite{apple2026rag}.
MobileRAG reduces vector-search and context-processing costs on mobile devices, and MUSE coordinates interactive vector queries with background index maintenance on mobile SoCs~\cite{park2025mobilerag,zhao2026muse}.
FAISS and HNSW provide established approximate-nearest-neighbor mechanisms for this retrieval kernel~\cite{johnson2021faiss,malkov2020hnsw}.
These systems motivate the retrieval workload studied here.

\paragraph{Agentic CPU--GPU Scheduling.}
Recent studies characterize CPU-heavy tool execution, map heterogeneous tools across CPUs and GPUs, or co-schedule multi-turn LLM--tool workflows~\cite{raj2025cpu,lu2026agentic,wang2026mars}.
They optimize end-to-end heterogeneous workflows, primarily on server platforms.
\system compares fixed, phase-aligned, and time-matched phase-blind admission of pending CPU retrieval while measuring TTFT and TPOT on unified memory.
Evaluating end-to-end agent dependencies and waiting times remains future work.

\end{document}